\documentclass{templates/arxiv/fairmeta}

\usepackage{amsmath}
\usepackage{amssymb}
\usepackage{mathtools}
\usepackage{bm}
\usepackage{algorithm}
\usepackage{algpseudocode}
\usepackage{paralist}
\usepackage[most]{tcolorbox}
\usepackage[dvipsnames,table,xcdraw]{xcolor}
\usepackage{booktabs}
\usepackage{array}
\usepackage{tabularx}
\usepackage{ragged2e}
\usepackage{makecell}
\usepackage{multirow}
\usepackage{tikz}
\usepackage{pgfplots}
\usepackage{wrapfig}
\usetikzlibrary{pgfplots.groupplots, matrix}
\pgfplotsset{compat=1.18}
\usepackage{enumitem}
\usepackage{comment}
\usepackage{graphicx}
\usepackage{multirow}
\usepackage{stackengine}
\usepackage{colortbl} % makes row coloring work better
\usepackage{anyfontsize}

\definecolor{purple}{HTML}{c994c7}
\definecolor{navyblue}{RGB}{30,130,255}
\definecolor{citecolor}{RGB}{30,130,255}
\definecolor{lightgray}{gray}{0.9}
\definecolor{blanchedalmond}{rgb}{1.0, 0.92, 0.8}
\definecolor{cerise}{rgb}{0.871, 0.192, 0.388}

\definecolor{TaskBG}{HTML}{EFE6FF}        % soft lavender (Task)
\definecolor{StateBG}{HTML}{F5F5F7}       % light gray (Current State) — unchanged
\definecolor{ExpertBG}{HTML}{EAF7EA}      % light green (Expert Action) — unchanged
\definecolor{IWMBG}{HTML}{FDECF3}         % light pink (Implicit World Modeling) — unchanged
\definecolor{SRBG}{HTML}{E6F2FF}          % light blue (Self-Reflection)

\newtcolorbox{trainingexample}[2][]{%
  enhanced, breakable, colframe=black!12, colback=white, boxrule=2.5pt,
  arc=2pt, left=0pt, right=0pt, top=0pt, bottom=0pt,
  title={#2}, fonttitle=\bfseries, coltitle=black, #1}

\newcolumntype{L}[1]{>{\RaggedRight\arraybackslash}p{#1}} % fixed-width paragraph
\newcolumntype{Y}{>{\RaggedRight\arraybackslash}X}        % stretch paragraph

\title{VideoResearcher: Self-Improving Tool Design for Long-Video Understanding}

\author[1,\star]{Dingqiang Ye}
\author[\star,\dagger]{Dongdi Zhao}
\author{Kaishen Wang}
\author{Qingqiao Hu}
\author{Jingchen Sun}
\author{Yijun Liang}
\author{Yuqi Jia}
\author{Yiqiao~Huang}
\author{Yunjie Tian}
\author{Jiaxing Zhang}
\author[1]{Chuanyang Jin}
\author[1]{Ke Zhang}
\author[1]{Vishal M. Patel}
\author[\ddagger]{Di Fu}

\affiliation[1]{Johns Hopkins University}
\contribution[\star]{Equal Contribution}
\contribution[\dagger]{Team Lead}
\contribution[\ddagger]{Corresponding Author}

\abstract{Video agents have made substantial progress in long-video understanding.
Yet effective video-agent systems require costly, time-consuming manual design and trial and error.
Current self-improvement methods either refine low-impact prompts, recombine predefined micro-tools, or struggle with convergence in harness optimization.
To bridge this gap, we target high-impact video-tool with \textbf{VideoResearcher}, a training-free multi-agent framework that autonomously designs, tests, and refines tools for video understanding, like a human researcher.
VideoResearcher operates through dual Solving and Evolving loops: it analyzes tool-use trajectories to identify capability gaps, coordinates specialized agents to develop and validate executable tools, and reuses evolved tools to strengthen evidence acquisition in subsequent video reasoning.
Through iterative tool refinement and validation, it progressively strengthens evidence acquisition without updating model parameters. 
VideoResearcher achieves state-of-the-art performance among self-improving agents and approaches the human-designed upper bound, demonstrating a training-free paradigm for long-video understanding that expands agent capabilities through autonomous tool development while reducing costly manual engineering.
}

\correspondence{Di Fu}

\begin{document}

\maketitle

% Keep large figures and tables at the top, with one float per page.
\setcounter{topnumber}{1}
\setcounter{dbltopnumber}{1}
\setcounter{totalnumber}{1}
\raggedbottom
% Keep paragraphs flowing across pages without isolated first or last lines.
\clubpenalty=10000
\widowpenalty=10000
\section{Introduction}
\label{sec:introduction}

Video understanding connects visual perception with language reasoning, supporting applications from interactive assistants to embodied systems.
Advances in large multimodal models have expanded these capabilities, enabling increasingly effective video dialogue and question answering~\citep{li2023videochat,zohar2025apollo}.
However, long videos introduce a harder requirement: models must locate relevant events and connect evidence scattered across time~\citep{wu2024longvideobench}.
Video agents largely address this challenge through iterative planning, tool use, and evidence gathering, allowing them to decide what to inspect next rather than answer from a single set of observations.
Their results show effective agentic reasoning for long-video understanding~\citep{wang2024videoagent,lin2026videoseek}.

\begin{figure}[t]
    \centering
    \includegraphics[width=\textwidth]{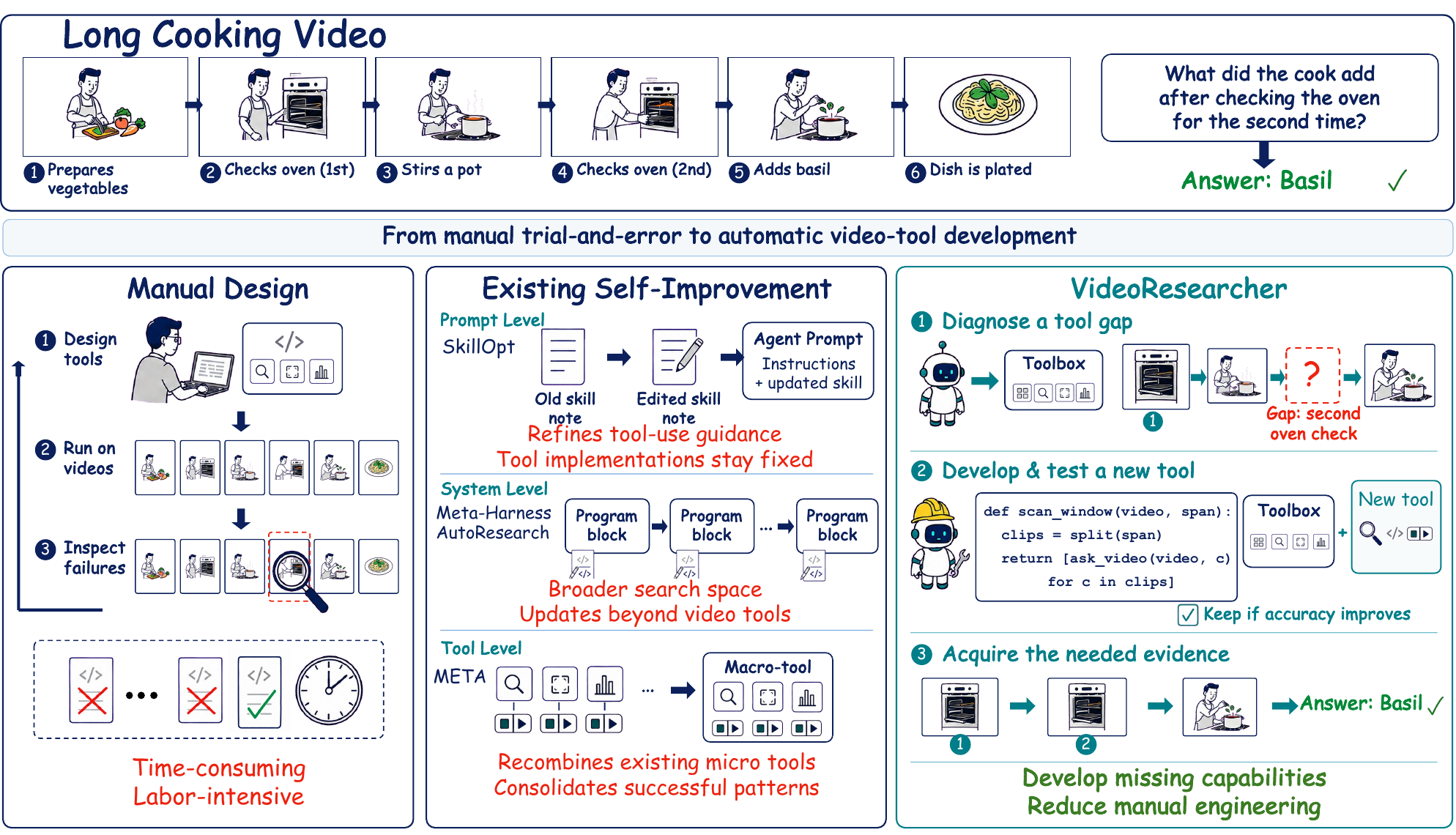}
    \caption{Motivation of VideoResearcher.
    Video-tool development remains labor-intensive.
    Existing self-improvement falls short: prompts have limited impact, system-wide search wastes most effort, and recombination remains bounded by predefined capabilities.
    VideoResearcher targets high-impact video-tool updates while avoiding costly system optimization.}
    \label{fig:overview}
\end{figure}

Existing video agents improve evidence acquisition through purpose-built tools.
VideoAgent~\citep{wang2024videoagent} uses vision-language models and retrieval tools to iteratively gather question-relevant information; DVD~\citep{zhang2025dvd} autonomously selects search tools over a multi-granular video database; and VideoSeek~\citep{lin2026videoseek} progressively localizes critical evidence through overview, skim, and focus tools.
These approaches demonstrate the value of tool-assisted reasoning, but developing effective toolkits requires researchers to identify needed operations, diagnose tool limitations, and validate improvements to video reasoning.
As illustrated on the left of Fig.~\ref{fig:overview}, researchers repeatedly execute the system, inspect trajectories, and revise tools, manually translating execution feedback into implementation changes.
Automating this process could reduce development effort and enable toolkits to improve through use.

A growing body of work seeks to automate this development process through agent self-improvement.
In general-purpose tasks, SkillOpt~\citep{yang2026skillopt} refines tool-use guidance, Meta-Harness~\citep{lee2026metaharness} optimizes whole agent-system code, and AutoResearch~\citep{karpathy2026autoresearch} automates code experimentation for model training.
For long-video understanding, META~\citep{huang2026meta} composes successful execution patterns into reusable macro-tools and extracts usage constraints from failures.
These efforts demonstrate the potential to automate parts of the development cycle, but their updates may offer limited gains in long-video understanding.
Better tool-use instructions~\citep{yang2026skillopt} or new combinations of predefined tools~\citep{huang2026meta} cannot directly address deficiencies in the underlying tools.
Changing the entire system~\citep{lee2026metaharness,karpathy2026autoresearch} allows broader exploration, but may largely waste effort on unrelated components.
This raises a question: how can we enable more efficient self-improvement for long-video understanding?

To address this question, we introduce \textbf{VideoResearcher}, a training-free multi-agent framework for targeted video-tool development.
\textit{Our key insight is that tool-level self-improvement addresses evidence-acquisition deficiencies more efficiently than instruction refinement or tool recombination, avoiding system-level redesign.}
Our framework consists of two coupled loops: a Solving loop that gathers execution experience and an Evolving loop that uses this experience to improve video tools.
The Solving loop separates reasoning, tool execution, and video perception, allowing the agent to actively gather evidence through VLM-based tools.
The Worker and perception model remain fixed, making the intermediate toolbox the sole target of self-improvement.
Within this scope, the Evolving loop coordinates specialized agents to develop and validate tool changes based on observed failures.
The two loops connect problem solving with tool development, enabling targeted improvements without modifying the surrounding agent architecture.

Extensive experiments on LVBench, LongVideoBench, and Video-MME evaluate the effectiveness of VideoResearcher.
VideoResearcher outperforms the evaluated same-base self-improvement methods in overall accuracy on all three benchmarks, raising the base agent's average accuracy from 72.1\% to 74.5\%.
These results demonstrate effective targeted tool development for long-video understanding.

Our contributions are threefold:
\begin{itemize}[leftmargin=*,topsep=2pt,itemsep=1pt]
    \item We introduce VideoResearcher, a training-free multi-agent framework that automates video-tool development for long-video understanding, reducing reliance on manual engineering.
    \item We propose failure-driven tool synthesis, where specialized agents translate evidence gaps into tool specifications and executable code, retaining updates that improve answer accuracy.
    \item We demonstrate state-of-the-art performance among evaluated self-improving agents on three long-video benchmarks and show how autonomous tool development strengthens evidence acquisition.
\end{itemize}

\section{Related Work}

\subsection{Long-Video Understanding}

Long-video understanding depends on both representing extended videos and finding relevant evidence.
Video-language models benefit from unified visual representations~\citep{li2024llava,wang2024qwen2,bai2025qwen2} and richer captioning and instruction data~\citep{chen2024sharegpt4video,zhang2025video}.
Long-video methods extend this foundation through hour-scale training~\citep{xue2024longvila,lin2025unleashing}, context transfer~\citep{zhang2024long,wei2025visual}, and efficient visual representations~\citep{shen2024longvu,shu2025video,xu2025slowfast,ye2024mplug,wang2025internvideo2}.
Pooling and token compression reduce redundant inputs~\citep{xu2024pllava,tao2025dycoke,yang2025pvc}, while reinforcement learning and self-distillation improve temporal reasoning and evidence use~\citep{feng2025video,wang2026look}.
Visual Contrastive Self-Distillation derives on-policy training targets by contrasting predictions conditioned on original and content-erased images~\citep{liang2026vcsd}.
Benchmarks such as HourVideo and Video-Holmes emphasize that longer context alone is insufficient: models must also recall events and connect scattered clues~\citep{chandrasegaran2024hourvideo,cheng2025video}.

Retrieval and memory make this evidence easier to access.
Methods organize video into searchable documents, temporal trees, or dynamic memories~\citep{ma2025drvideo,wang2025videotree,pang2025mrvideo,luo2025video,zuo2025videolucy}; others retain past observations~\citep{he2024ma,fan2024videoagent}, retrieve from video collections~\citep{jeong2025videorag}, or query hierarchical graph memories~\citep{chen2026memdreamer}.
Agents further control evidence acquisition during reasoning: VideoAgent~\citep{wang2024videoagent} retrieves observations, DVD~\citep{zhang2025dvd} searches a multi-granular database, and VideoSeek~\citep{lin2026videoseek} moves from overview to focused inspection.
Related agents adapt temporal search~\citep{yang2025vca,yin2025videoarm,tian2025ego}, observation granularity~\citep{wang2026avp,li2026lenswalk}, or multi-agent coordination~\citep{xu2026a4vl,yan2026symphony,chen2026videochatm1}.
These methods improve evidence access, but their available operations are generally specified before execution.
VideoResearcher automates this cycle: diagnosing missing capabilities, implementing tool changes, and testing them against the current toolbox.

\subsection{Agent Evolution}

Agent evolution improves guidance, system code, or reusable capabilities through execution feedback.
Self-Refine revises outputs through self-feedback~\citep{madaan2023selfrefine}; Reflexion retains verbal lessons across attempts~\citep{shinn2023reflexion}.
DSPy, TextGrad, and GEPA optimize language-model programs through task evaluations or textual feedback~\citep{khattab2024dspy,yuksekgonul2025textgrad,agrawal2025gepa}.
SkillOpt refines reusable tool-use instructions~\citep{yang2026skillopt}; Trace2Skill extracts trajectory lessons~\citep{ni2026trace2skill}, while EvoSkill and CoEvoSkills develop skills through failure analysis or co-evolving verification~\citep{alzubi2026evoskill,zhang2026coevoskills}.
Skills may contain instructions, code, or both, requiring evaluation in use~\citep{jiang2026sok,li2026skillsbench}.\looseness=-1

Broader approaches optimize the agent system itself.
Meta-Harness and VeRO improve the code surrounding model calls~\citep{lee2026metaharness,ursekar2026vero}; SkillOpt-Lite and EvoTest also refine agent behavior from trajectories~\citep{shen2026skilloptlite,he2025evotest}.
Architecture and workflow search modify roles, coordination, and execution structure~\citep{hu2025automated,zhang2025aflowautomatingagenticworkflow,song2026abstral,zhang2026dgm}.
Related program-evolution systems evaluate code changes in mathematical, algorithmic, or model-training tasks~\citep{romera2024mathematical,novikov2025alphaevolve,openevolve,zelikman2024stop,karpathy2026autoresearch}.
These methods offer broad flexibility; VideoResearcher focuses on tools that acquire video evidence.

Executable tool learning is the closest line of work.
LATM and CREATOR separate tool creation from use~\citep{cai2024latm,qian2023creator}; Voyager accumulates code skills from environment feedback~\citep{wang2024voyager}, and AgentOptimizer improves callable functions while freezing the language model~\citep{zhang2024agentoptimizer}.
ToolMaker and CASCADE construct tools or scientific skills from external resources~\citep{wolflein2025toolmaker,huang2025cascade}, while experience reuse can develop tools for multimodal reasoning~\citep{shen2026uct}.
For long videos, META consolidates recurring micro-tool trajectories into macro-tools and derives usage priors from failures~\citep{huang2026meta}.
VideoResearcher targets failure-directed video-tool development: structured diagnosis identifies an evidence gap, specialized agents implement a candidate, and development accuracy governs acceptance with fixed Worker and perception models.

\section{Method}
\label{sec:method}

\subsection{Overview}
\begin{figure}[t]
    \centering
    \includegraphics[width=\textwidth]{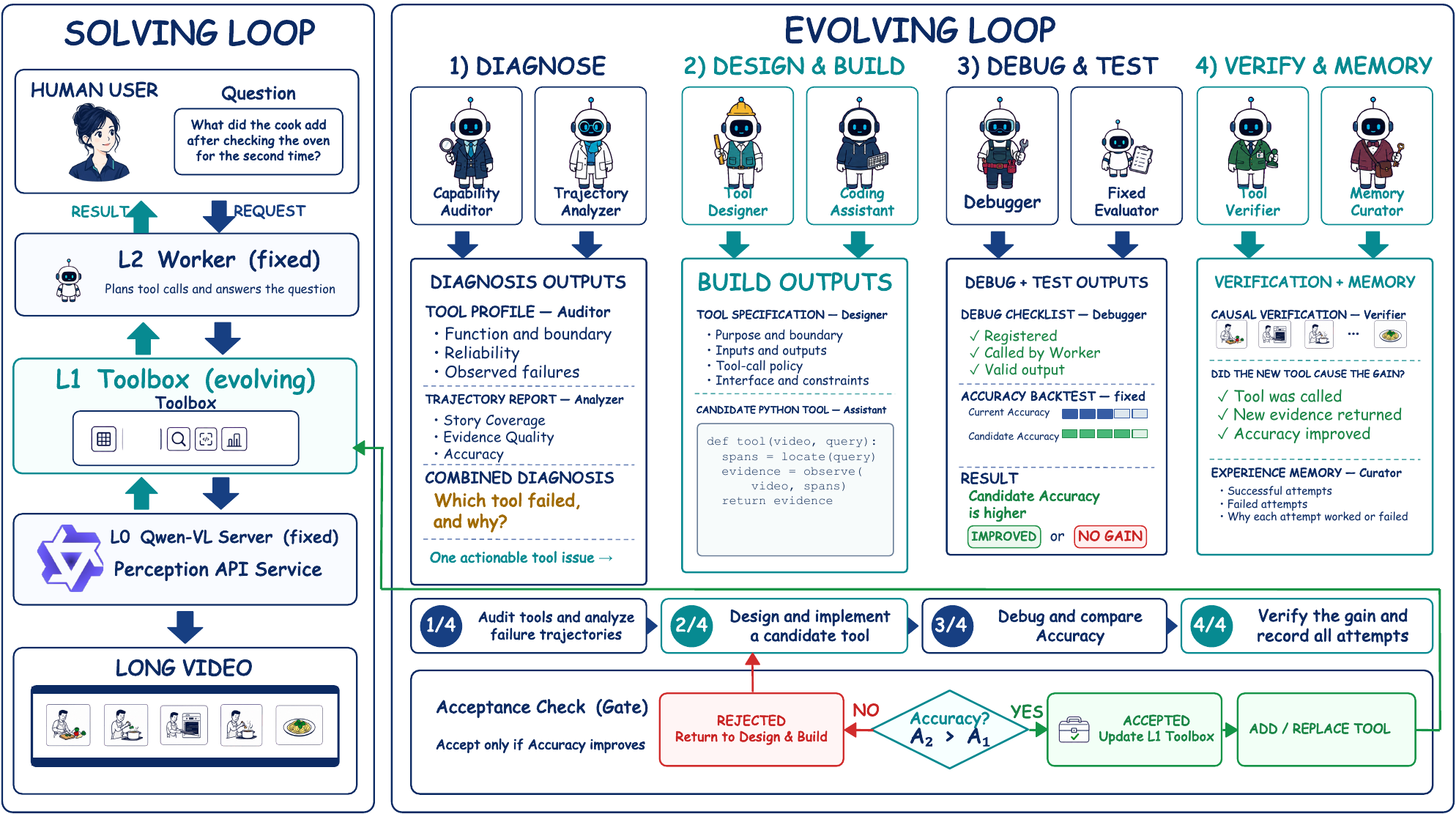}
    \caption{Overview of VideoResearcher.
    The Solving Loop connects a fixed Worker to a fixed perception model through an evolving toolbox.
    The Evolving Loop diagnoses trajectory failures, develops and tests executable tools, inspects their effects, and updates the toolbox.}
    \label{fig:method-overview}
\end{figure}

\begin{samepage}
VideoResearcher couples Solving and Evolving Loops to improve video tools (Fig.~\ref{fig:method-overview}).
The Solving Loop answers questions and records trajectories with the current toolbox; the Evolving Loop diagnoses failures, develops and tests executable tools, and retains updates that improve development accuracy.
The Worker and perception model stay fixed; evolution changes evidence acquisition without parameter updates or system redesign.\looseness=-1\par
\end{samepage}

\begin{samepage}
For a task $x_i=(V_i,q_i)$ with video $V_i$, question $q_i$, and reference answer $y_i$, the system at evolution round $r$ is
\begin{equation}
    \Pi^{(r)}=\bigl(\pi_{\mathrm{W}},\mathcal{T}^{(r)},P\bigr),
    \qquad
    \mathcal{T}^{(0)}=\{\texttt{ask\_video}\},
    \label{eq:architecture}
\end{equation}
where $\pi_{\mathrm{W}}$ is the Worker, $P$ is the perception model, and $\mathcal{T}^{(r)}$ is the toolbox, initially containing only the general-purpose \texttt{ask\_video} tool.
Fixing $\pi_{\mathrm{W}}$ and $P$ isolates toolbox changes across rounds.\par
\end{samepage}

On a fixed development set $\mathcal{D}_{\mathrm{evo}}$, the loops collect experience and update the toolbox:
\begin{equation}
\begin{aligned}
 (\hat y_i^{(r)},\tau_i^{(r)})
    &=\operatorname{Solve}(x_i;\Pi^{(r)}),\\
 \mathcal{E}^{(r)}
    &=\{(x_i,y_i,\hat y_i^{(r)},\tau_i^{(r)})\}_{(x_i,y_i)\in\mathcal{D}_{\mathrm{evo}}},\\
 \mathcal{T}^{(r+1)}
    &=\operatorname{Evolve}\!\left(
      \mathcal{T}^{(r)},\mathcal{E}^{(r)};A_{\mathrm{val}}
      \right),
\end{aligned}
\label{eq:dual-loop}
\end{equation}
Experience $\mathcal{E}^{(r)}$ comprises tasks, reference answers, predictions $\hat y_i^{(r)}$, and trajectories $\tau_i^{(r)}$.
The Evolving Loop develops candidates from this experience, accepting only those that improve development-set accuracy $A_{\mathrm{val}}$.
Sections~\ref{sec:solving} and~\ref{sec:evolving} detail trajectory collection and tool development.

\par\penalty0
\begin{samepage}
\subsection{Solving: Hierarchical Tool Use}
\label{sec:solving}
In the Solving Loop (Fig.~\ref{fig:method-overview}), the Worker (L2) requests evidence, tools (L1) acquire it, and the perception model (L0) inspects video.
It interleaves reasoning and action~\citep{yao2023react} with active evidence seeking~\citep{wang2026avp}, evolving observation procedures with both models fixed.\looseness=-1\par
\end{samepage}

\par\penalty0
\begin{samepage}
\paragraph{Worker-guided reasoning.}
The Worker accesses video only through tools.
At step $j$, it uses the question and interaction history $h_{i,<j}^{(r)}$ to select a tool $g_{ij}^{(r)}\in\mathcal{T}^{(r)}$ and arguments $a_{ij}^{(r)}$:
\begin{equation}
 (g_{ij}^{(r)},a_{ij}^{(r)})
 =\pi_{\mathrm{W}}\!\left(q_i,h_{i,<j}^{(r)};\mathcal{T}^{(r)}\right).
 \label{eq:worker-policy}
\end{equation}
The Worker appends each observation to its history, then requests more evidence or produces $\hat y_i^{(r)}$.\looseness=-1\par
\end{samepage}

\par\penalty0
\begin{samepage}
\paragraph{Tool-mediated perception.}
Tools translate evidence requests into perception calls.
The initial tool exposes the perception model directly:
\begin{equation}
 \texttt{ask\_video}(V_i;p,s,e,\rho)
 =P(V_i,p,s,e,\rho).
 \label{eq:ask-video}
\end{equation}
Here, $p$ is the visual query passed as \texttt{question}, $s$ and $e$ are optional start and end times in seconds, and $\rho$ is the sampling rate in frames per second (FPS).
The tool defaults to the full video at 2 FPS and returns a textual observation.
Evolved tools use the same perception interface but may define their own arguments, combine targeted calls, and aggregate observations into reusable procedures.\par
\end{samepage}

\paragraph{Trajectory collection.}
Each trajectory preserves the interactions that produced an answer.
For tool observation $o_{ij}^{(r)}$, we record
\begin{equation}
\begin{aligned}
 o_{ij}^{(r)}
   &=g_{ij}^{(r)}\!\left(V_i,a_{ij}^{(r)};P\right),
 &c_{ij}^{(r)}
   &=\left(g_{ij}^{(r)},a_{ij}^{(r)},o_{ij}^{(r)}\right),\\
 \tau_i^{(r)}
   &=\left(c_{i1}^{(r)},\ldots,c_{iM_i}^{(r)},\hat y_i^{(r)}\right),
\end{aligned}
\label{eq:trajectory}
\end{equation}
where $M_i$ is the number of tool calls.
The record includes underlying perception requests, giving the Evolving Loop the tool choices, arguments, and evidence needed for diagnosis.

\subsection{Evolving: Multi-Agent Tool Development}
\label{sec:evolving}
The Evolving Loop assigns diagnosis, implementation, testing, and verification to specialized agents linked by structured reports (Fig.~\ref{fig:method-overview}):
\begin{equation}
 \mathcal{E}^{(r)}
 \xrightarrow[\mathcal{T}^{(r)}]{\text{diagnose}} d^{(r)}
 \xrightarrow[\mathcal{T}^{(r)},\,\mathcal{M}^{(r)}]{\text{design and build}} \widetilde g^{(r)}
 \xrightarrow{\text{debug and test}} b^{(r)}
 \xrightarrow[\mathcal{M}^{(r)}]{\text{verify and remember}}
 \bigl(\mathcal{T}^{(r+1)},\mathcal{M}^{(r+1)}\bigr),
 \label{eq:evolving-workflow}
\end{equation}
Here, $d^{(r)}$ is a diagnosis, $\widetilde g^{(r)}$ a candidate tool, $b^{(r)}$ a comparison against the current toolbox, and $\mathcal{M}^{(r)}$ the accumulated development experience.
The lower labels specify inputs: diagnosis inspects the current toolbox, development uses it as the implementation base, and stored experience guides design and coding.
Each candidate is thus linked to an observed failure and a recorded outcome.

\paragraph{1) Diagnose.}
Diagnosis narrows execution failures to one tool-development target.
The Capability Auditor produces a \emph{Tool Profile} describing each tool's function, boundaries, reliability, and limitations.
The Trajectory Analyzer combines this profile with trajectories to produce a \emph{Trajectory Report} of recurring failures and answer outcomes.
Its diagnostic criteria are \emph{Story Coverage}, whether observations cover relevant events, and \emph{Evidence Quality}, whether they provide useful evidence for answering the question.
These criteria guide diagnosis rather than serve as acceptance objectives.
The reports specify $d^{(r)}$: the failure, its conditions, and the missing capability.\looseness=-1

\paragraph{2) Design and build.}
Development turns $d^{(r)}$ into a specification and executable code.
Using the \emph{Tool Profile} and memory $\mathcal{M}^{(r)}$, the Tool Designer specifies the tool's purpose, input--output schema, boundaries, perception-call policy, and difference from existing tools in a \emph{Tool Specification}.
The Coding Assistant implements it as a \emph{Candidate Python Tool} $\widetilde g^{(r)}$ in an isolated toolbox.
Changes may affect perception prompts, temporal decomposition, inspected intervals, frame selection, and output aggregation.
The tool can be registered, invoked, and evaluated by the fixed Worker (Table~\ref{tab:tool-evolution}).

% Declare the results float early enough for the next page top.
% Compact main-results table without Perception and Worker columns.
% Switch variants at the results-float anchor in sections/003-method.tex.
\begin{table*}[t]
\centering
\small
\definecolor{deltagain}{HTML}{008A45}
\definecolor{deltaloss}{HTML}{D63B32}
\newcommand{\resultdelta}[1]{\hspace{0.10em}{\normalfont\scriptsize\ifdim #1pt>0pt\color{deltagain}\else\ifdim #1pt<0pt\color{deltaloss}\else\color{black!55}\fi\fi\ensuremath{(#1)}}}
\setlength{\tabcolsep}{4.4pt}
\renewcommand{\arraystretch}{1.15}
\caption{
Accuracy (\%) on three representative long-video benchmarks.
$^{*}$ marks methods evolved from the same Base, VideoResearcher (w/o Evolution); parentheses show changes from it.
Average is computed over the three Overall scores.
The best and second-best results are bolded and underlined.
}
\label{tab:main-results}
\resizebox{\textwidth}{!}{%
\begin{tabular}{lccccccc}
\toprule
\multirow{2}{*}{\textbf{Method}}
& \multirow{2}{*}{\textbf{Venue}}
& \textbf{LVBench}
& \multicolumn{2}{c}{\textbf{LongVideoBench}}
& \multicolumn{2}{c}{\textbf{Video-MME}}
& \textbf{Average} \\
\cmidrule(lr){3-3}\cmidrule(lr){4-5}\cmidrule(lr){6-7}\cmidrule(l){8-8}
& & \textbf{Overall}
& \textbf{Overall} & \textbf{Long}
& \textbf{Overall} & \textbf{Long}
& \textbf{Overall} \\
\midrule
\rowcolor{blue!5}
\multicolumn{8}{l}{\textit{Human-engineered agentic video systems}}\\
VideoAgent~\citep{wang2024videoagent} & ECCV'24 & 29.3 & -- & -- & -- & 46.4 & -- \\
VideoTree~\citep{wang2025videotree} & CVPR'25 & 28.8 & -- & -- & 60.6 & 54.2 & -- \\
% DrVideo~\citep{ma2025drvideo} & CVPR'25 & -- & -- & -- & -- & 51.7 & -- \\
% VCA~\citep{yang2025vca} & ICCV'25 & 41.3 & -- & -- & -- & -- & -- \\
% MR.~Video~\citep{pang2025mrvideo} & NeurIPS'25 & 60.8 & -- & 61.6 & -- & 61.8 & -- \\
% LVAgent~\citep{chen2025lvagent} & ICCV'25 & -- & 80.0 & -- & 81.7 & 74.3 & -- \\
VideoLucy~\citep{zuo2025videolucy} & NeurIPS'25 & -- & 58.8 & -- & 72.5 & 66.8 & -- \\
DVD~\citep{zhang2025dvd} & NeurIPS'25 & \textbf{74.2} & 71.6 & 68.6 & -- & 67.3 & -- \\
VideoSeek~\citep{lin2026videoseek} & CVPR'26 & \underline{68.4} & -- & \textbf{73.5} & -- & 70.1 & -- \\
VideoARM~\citep{yin2025videoarm} & CVPR'26 & -- & 73.7 & 69.2 & \underline{80.1} & \textbf{75.3} & -- \\
% A4VL~\citep{xu2026a4vl} & CVPR'26 & -- & 72.2 & -- & 77.2 & 68.3 & -- \\
% LensWalk~\citep{li2026lenswalk} & CVPR'26 & 68.6 & -- & 70.6 & -- & 71.4 & -- \\
% Symphony~\citep{yan2026symphony} & CVPR'26 & -- & 77.1 & -- & -- & -- & -- \\
% VideoChat-M1~\citep{chen2026videochatm1} & CVPR'26 & -- & 82.3 & -- & 83.2 & -- & -- \\
% AVP~\citep{wang2026avp} & CVPR-F'26 & 63.8 & 70.2 & 65.5 & 81.2 & 76.7 & 71.7 \\
VideoResearcher (w/o Evolution) & Ours & 66.9 & 73.2 & 70.6 & 76.3 & 70.6 & 72.1 \\
\midrule
\rowcolor{blue!5}
\multicolumn{8}{l}{\textit{Self-improving agentic video systems}}\\
META ~\citep{huang2026meta} & CVPR'26 & -- & 63.6 & -- & 77.7 & -- & -- \\
SkillOpt$^{*}$~\citep{yang2026skillopt} & arXiv'26 & 64.8\resultdelta{-2.1} & 66.5\resultdelta{-6.7} & 63.8\resultdelta{-6.8} & 78.1\resultdelta{+1.8} & 74.1\resultdelta{+3.5} & 69.8\resultdelta{-2.3} \\
Meta-Harness$^{*}$~\citep{lee2026metaharness} & arXiv'26 & 67.1\resultdelta{+0.2} & \underline{74.9}\resultdelta{+1.7} & 70.9\resultdelta{+0.3} & 78.4\resultdelta{+2.1} & 73.6\resultdelta{+3.0} & \underline{73.5}\resultdelta{+1.4} \\
AutoResearch$^{*}$~\citep{karpathy2026autoresearch} & GitHub'26 & 66.8\resultdelta{-0.1} & 73.0\resultdelta{-0.2} & \underline{73.4}\resultdelta{+2.8} & 75.2\resultdelta{-1.1} & 73.3\resultdelta{+2.7} & 71.7\resultdelta{-0.4} \\
% VideoResearcher (Qwen3-VL-8B)
% & Ours
% & 66.8
% & \(\approx 77.4\) & \(\approx 73.6\)
% & \(\approx 79.7\) & 77.1
% & \(\approx 74.6\) \\
% \rowcolor{blue!3}
VideoResearcher
& Ours
& \underline{68.4}\resultdelta{+1.5}
& \textbf{75.0}\resultdelta{+1.8} & 71.6\resultdelta{+1.0}
& \textbf{80.2}\resultdelta{+3.9} & \underline{74.7}\resultdelta{+4.1}
& \textbf{74.5}\resultdelta{+2.4} \\
% VideoResearcher (Qwen3.5-9B)
% & Ours
% & 68.0
% & - & -
% & 80.0 & 73.9
% & -- \\
\bottomrule
\end{tabular}
}
\end{table*}

\paragraph{3) Debug and test.}
Testing checks both execution and answer accuracy.
The Debugger runs smoke tests and records a \emph{Debug Checklist} covering registration, execution, Worker invocation, schema compliance, and output validity.
Failures return to the Coding Assistant for a bounded repair loop.
After these checks pass, a fixed Evaluator compares the candidate and current toolboxes on the same development examples, using the same Worker, perception model, inference settings, and random seed.
This \emph{Accuracy Backtest} $b^{(r)}$ uses
\begin{equation}
 A_{\mathrm{val}}(\mathcal{T})=
 \frac{1}{|\mathcal{D}_{\mathrm{evo}}|}
 \sum_{(x_i,y_i)\in\mathcal{D}_{\mathrm{evo}}}
 \mathbb{I}\!\left[\hat y_i(\mathcal{T})=y_i\right].
 \label{eq:validation-accuracy}
\end{equation}
Diagnosis and selection use this development set; final benchmarks remain separate (Sec.~\ref{sec:experiments}).

\paragraph{4) Verify and remember.}
Verification examines how a candidate changed the evidence and preserves the result for future development.
The Tool Verifier's \emph{Causal Verification Report} records candidate invocation, new observations, and their relation to corrected or regressed answers.
This is a trajectory-based explanation of the backtest, not an independent causal experiment.
It does not override the sole acceptance criterion, an accuracy increase:
\begin{equation}
\mathcal{T}^{(r+1)}=
\begin{cases}
\widetilde{\mathcal{T}}^{(r+1)},
& A_{\mathrm{val}}(\widetilde{\mathcal{T}}^{(r+1)})
  > A_{\mathrm{val}}(\mathcal{T}^{(r)}),\\
\mathcal{T}^{(r)},&\text{otherwise},
\end{cases}
\label{eq:acceptance}
\end{equation}
where $\widetilde{\mathcal{T}}^{(r+1)}=\operatorname{Update}(\mathcal{T}^{(r)},\widetilde g^{(r)})$ is the isolated candidate toolbox.
The Memory Curator appends the attempt's status, design, result, and lesson to \emph{Experience Memory}, producing $\mathcal{M}^{(r+1)}$.
Accepted tools enter the next Solving Loop; rejected attempts inform future design and coding.
This links failures to tested tool updates.\looseness=-1

\section{Experiments}
\label{sec:experiments}

\begin{table}[t]
\centering
\small
\caption{
Analysis of framework components, model choices, and tool evolution.
Panel (d) breaks down LVBench by question type.
ER, EU, KIR, TG, Rea, and Sum denote entity recognition, event understanding, key-information retrieval, temporal grounding, reasoning, and summarization.
}
\label{tab:ablation}
\begin{minipage}[t]{0.40\linewidth}
\centering
\footnotesize\textbf{(a) Framework components}\par\vspace{2pt}
\scriptsize
\setlength{\tabcolsep}{1pt}
\renewcommand{\arraystretch}{1.06}
\begin{tabular*}{\linewidth}{@{\extracolsep{\fill}}cccc@{}}
\toprule
\multirow{2}{*}{\textbf{Worker}} &
\multirow{2}{*}{\textbf{Evolution}} &
\multicolumn{2}{c}{\textbf{Perception}} \\
\cmidrule(lr){3-4}
& & \textbf{Qwen3-VL-8B} & \textbf{Qwen3.5-4B} \\
\midrule
$\times$ & $\times$ & 58.0 & 57.6 \\
$\checkmark$ & $\times$ & 63.6 & 66.9 \\
$\checkmark$ & $\checkmark$ & \textbf{66.8} & \textbf{68.4} \\
\bottomrule
\end{tabular*}
\end{minipage}
\hfill
\begin{minipage}[t]{0.58\linewidth}
\centering
\footnotesize\textbf{(b) Overall accuracy across benchmarks}\par\vspace{2pt}
\scriptsize
\setlength{\tabcolsep}{1pt}
\renewcommand{\arraystretch}{1.06}
\begin{tabular*}{\linewidth}{@{\extracolsep{\fill}}lccc@{}}
\toprule
\multirow{2}{*}{\textbf{Method}} & \multicolumn{3}{c}{\textbf{Benchmark}} \\
\cmidrule(lr){2-4}
& \textbf{LVBench} & \textbf{Video-MME} & \textbf{LongVideoBench} \\
\midrule
VideoResearcher (w/o Evolution) & 66.9 & 76.3 & 73.2 \\
VideoResearcher (Round 2) & \textbf{68.9} & 80.0 & \textbf{75.5} \\
VideoResearcher (Round 7) & 68.4 & \textbf{80.2} & 75.0 \\
\bottomrule
\end{tabular*}
\end{minipage}
\par\vspace{7pt}
\begin{minipage}[t]{0.40\linewidth}
\vspace{0pt}
\centering
\footnotesize\textbf{(c) Worker and perception models}\par\vspace{2pt}
\scriptsize
\setlength{\tabcolsep}{1pt}
\renewcommand{\arraystretch}{1.06}
\begin{tabular*}{\linewidth}{@{\extracolsep{\fill}}lcc@{}}
\toprule
\multirow{2}{*}{\textbf{Worker}} & \multicolumn{2}{c}{\textbf{Perception}} \\
\cmidrule(lr){2-3}
& \textbf{Qwen3-VL-8B} & \textbf{Qwen3.5-4B} \\
\midrule
GPT-o4-mini & 45.0 & 45.6 \\
GPT-5.3-Codex & 44.3 & 50.5\\
GPT-5.6-SOL & \textbf{66.8} & \textbf{68.4} \\
\bottomrule
\end{tabular*}
\end{minipage}
\hfill
\begin{minipage}[t]{0.58\linewidth}
\vspace{0pt}
\centering
\footnotesize\textbf{(d) LVBench accuracy by question type}\par\vspace{2pt}
\scriptsize
\setlength{\tabcolsep}{1pt}
\renewcommand{\arraystretch}{1.06}
\begin{tabular*}{\linewidth}{@{\extracolsep{\fill}}lccccccc@{}}
\toprule
\multirow{2}{*}{\textbf{Method}} & \multicolumn{6}{c}{\textbf{Question type}} & \multirow{2}{*}{\textbf{Overall}} \\
\cmidrule(lr){2-7}
& \textbf{ER} & \textbf{EU} & \textbf{KIR} & \textbf{TG} & \textbf{Rea} & \textbf{Sum} & \\
\midrule
VideoResearcher (w/o Evolution) & 66.2 & 65.2 & \textbf{77.3} & 67.7 & 61.2 & 60.3 & 66.9 \\
VideoResearcher (Round 2) & 66.2 & \textbf{68.3} & 72.9 & 71.4 & \textbf{65.7} & \textbf{77.6} & \textbf{68.9} \\
VideoResearcher (Round 7) & \textbf{66.6} & 67.4 & 74.6 & \textbf{72.3} & 64.7 & 69.0 & 68.4 \\
\bottomrule
\end{tabular*}
\end{minipage}
\end{table}

\subsection{Experimental Setup}
\label{sec:experimental-setup}
\paragraph{Benchmarks.}
We evaluate on three long-video benchmarks.
LVBench~\citep{wang2025lvbench} comprises 103 videos (30 minutes--2 hours) and 1,549 multiple-choice questions across six categories.
LongVideoBench~\citep{wu2024longvideobench} contains 3,763 videos (up to 1 hour) and 6,678 questions for long-context retrieval and reasoning.
Video-MME~\citep{fu2025video} comprises 900 videos (254 hours) and 2,700 questions across six domains and three duration groups.
We report overall accuracy (Overall) on all three benchmarks and long-duration subset accuracy (Long) for LongVideoBench and Video-MME, using subtitles only for LongVideoBench.\par

\paragraph{Implementation details.}
Unless otherwise noted, GPT-5.6-SOL and Qwen3.5-4B in instruct mode serve as the fixed Worker and perception model, respectively; Evolving roles use the Codex SDK.
The Solving Loop allows at most 10 turns of tool use per question.
The initial \texttt{ask\_video} tool (Sec.~\ref{sec:solving}) uses one perception call and at most 2,048 sampled frames.
The Evolving Loop uses up to 20 rounds and 300 fixed ALLVB~\citep{tan2025allvb} development questions; videos average nearly two hours.
Each evolved tool may make at most five perception calls and sample at most 512 frames in total.
Only updates that strictly improve development accuracy are retained, and the toolbox is frozen for final testing on the three benchmarks, which are excluded from tool design and selection.
The Qwen service was deployed using eight NVIDIA B200 GPUs.\par

\paragraph{Baselines and controlled comparison.}
Table~\ref{tab:main-results} compares human-engineered and self-improving video agents.
SkillOpt~\citep{yang2026skillopt}, Meta-Harness~\citep{lee2026metaharness}, AutoResearch~\citep{karpathy2026autoresearch}, and VideoResearcher start from \emph{VideoResearcher (w/o Evolution)}, called the \emph{Base}, with the same Worker, perception model, inference limits, initial \texttt{ask\_video} tool, 300 development questions, and 20-round evolution budget.
This comparison varies the self-improvement strategy from a common starting agent.
Human-engineered agents and META~\citep{huang2026meta} use published configurations and scores under the closest matching benchmark settings; they provide external reference points rather than matched controls.\par

\begin{table}[t]
\centering
\caption{Tool-evolution outcomes and an accepted executable tool. Panel~(b) shows the Round-2 tool's contract, prompt, and core implementation; \texttt{Qwen} denotes the fixed perception model. Accuracy is measured on 300 development questions from ALLVB~\citep{tan2025allvb}.}
\label{tab:tool-evolution}
\vspace{2pt}

\providecommand{\ToolTableStyleBegin}{}
\providecommand{\ToolTableStyleEnd}{}
\ToolTableStyleBegin
\setlength{\parskip}{0pt}
\newsavebox{\ToolEvolutionLeftBox}
\newsavebox{\ToolEvolutionRightBox}
\newsavebox{\ToolEvolutionCodeBox}
\newlength{\ToolEvolutionLeftWidth}
\newlength{\ToolEvolutionRightWidth}
\newlength{\ToolEvolutionPanelHeight}
\newlength{\ToolEvolutionRowPad}
\setlength{\ToolEvolutionLeftWidth}{0.590\linewidth}
\setlength{\ToolEvolutionRightWidth}{0.390\linewidth}
\setlength{\ToolEvolutionRowPad}{2pt}

% Keep each tool's name, failure, and behavior together in a readable wide cell.
\newcommand{\ToolEvolutionInfo}[3]{%
  \vspace*{\ToolEvolutionRowPad}%
  \textbf{#1}\par
  {\color{black!65}\textit{Target:}} #2\par
  {\color{black!65}\textit{Function:}} #3%
  \par\vspace*{\ToolEvolutionRowPad}%
}
\newcommand{\ToolEvolutionTable}{%
  \fontsize{7.5}{8.7}\selectfont
  \setlength{\tabcolsep}{2.2pt}%
  \renewcommand{\arraystretch}{1.04}%
  \renewcommand{\tabularxcolumn}[1]{m{##1}}%
  \begin{tabularx}{\linewidth}{@{}>{\centering\arraybackslash}m{21pt}>{\raggedright\arraybackslash}X>{\centering\arraybackslash}m{45pt}@{}}
  \toprule
  \textbf{Round} & \textbf{Evolved tool} &
  \textbf{Dev. acc.}\par\textbf{Result} \\
  \midrule
  \rowcolor{blue!5}
  2 &
  \ToolEvolutionInfo{Neutral Scene Landmark Index}
    {Unreliable coarse event localization}
    {Build a five-part landmark index from about 512 frames} &
  $68.7\!\rightarrow\!75.0$\par
  \textcolor{ForestGreen}{\textbf{Accept}} \\
  3 &
  \ToolEvolutionInfo{Premise-free action anchor index}
    {Incomplete global action coverage}
    {Extract early, middle, and late action anchors from five segments} &
  $75.0\!\rightarrow\!70.7$\par
  \textcolor{BrickRed}{Reject} \\
  \rowcolor{blue!5}
  7 &
  \ToolEvolutionInfo{Candidate Window Atomic Scan}
    {Weak evidence near a candidate event}
    {Scan five 20-s clips at 5 fps for timestamped atomic events} &
  $75.0\!\rightarrow\!76.7$\par
  \textcolor{ForestGreen}{\textbf{Accept}} \\
  14 &
  \ToolEvolutionInfo{Appearance--object signature index}
    {Broken cross-scene entity links}
    {Link appearance, object, and interaction signatures across scenes} &
  $76.7\!\rightarrow\!69.0$\par
  \textcolor{BrickRed}{Reject} \\
  17 &
  \ToolEvolutionInfo{Single-context midpoint evidence reel}
    {Evidence fragmented across calls}
    {Inspect one midpoint window in a single context} &
  $76.7\!\rightarrow\!71.7$\par
  \textcolor{BrickRed}{Reject} \\
  \bottomrule
  \end{tabularx}%
}

% Capture the listing once. Adjacent strings preserve the complete prompt.
\begin{lrbox}{\ToolEvolutionCodeBox}
\begin{minipage}{\dimexpr\ToolEvolutionRightWidth-10pt\relax}
\begin{lstlisting}[
  language=Python,
  basicstyle=\ttfamily\fontsize{7}{8.1}\selectfont,
  keywordstyle=\color{NavyBlue},
  stringstyle=\color{BrickRed},
  commentstyle=\color{black!55}\itshape,
  showstringspaces=false,
  columns=fullflexible,
  keepspaces=true,
  breaklines=true,
  breakatwhitespace=false,
  aboveskip=0pt,
  belowskip=0pt,
  tabsize=2
]
def neutral_scene_landmark_index(video):
    prompt = (
      "Analyze the clip and summarize "
      "visible events in temporal order. "
      "Infer roles and relationships "
      "from visible evidence. "
      "Consider alternatives and "
      "uncertainty before summarizing."
    )
    T = video_duration(video)
    dt, fps, segments = T / 5, 512 / T, []
    for i in range(5):
        start, end = (
          i * dt, min((i + 1) * dt, T))
        desc = Qwen(
          video, start, end, fps, prompt)
        segments.append({
          "segment": i + 1, "start": start,
          "end": end, "description": desc})
    return {"segments": segments}
\end{lstlisting}
\end{minipage}
\end{lrbox}

\newcommand{\ToolEvolutionRightPanel}[1]{%
  \begin{tcolorbox}[
    enhanced,
    width=\linewidth,
    before skip=0pt, after skip=0pt,
    colback=gray!3, colframe=black!25,
    boxrule=0.4pt, boxsep=0pt, arc=1pt,
    left=4.6pt, right=4.6pt, top=4pt, bottom=4pt,
    grow to left by=0pt, grow to right by=0pt,
    valign=center,
    #1
  ]
  \raggedright\fontsize{7.5}{8.7}\selectfont
  \textbf{Tool contract}\par\vspace{2pt}
  Input: video path. Output: five ordered scene landmarks.
  Budget: five Qwen calls and approximately 512 sampled frames.
  \par\vspace{4pt}
  {\color{black!20}\hrule height 0.4pt}
  \vspace{4pt}
  \textbf{Evolved function}\par\vspace{3pt}
  \usebox{\ToolEvolutionCodeBox}%
  \end{tcolorbox}%
}

% Measure natural heights and distribute a left-panel shortfall across rows.
\begin{lrbox}{\ToolEvolutionLeftBox}
\begin{minipage}{\ToolEvolutionLeftWidth}
\ToolEvolutionTable
\end{minipage}
\end{lrbox}
\begin{lrbox}{\ToolEvolutionRightBox}
\begin{minipage}{\ToolEvolutionRightWidth}
\ToolEvolutionRightPanel{}
\end{minipage}
\end{lrbox}
\setlength{\ToolEvolutionPanelHeight}{\dimexpr\ht\ToolEvolutionLeftBox+\dp\ToolEvolutionLeftBox\relax}
\typeout{Tool evolution natural left: \the\ToolEvolutionPanelHeight}
\typeout{Tool evolution natural right: \the\dimexpr\ht\ToolEvolutionRightBox+\dp\ToolEvolutionRightBox\relax}
\ifdim\dimexpr\ht\ToolEvolutionRightBox+\dp\ToolEvolutionRightBox\relax>\ToolEvolutionPanelHeight
  \setlength{\ToolEvolutionRowPad}{\dimexpr
    \ToolEvolutionRowPad+
    (\ht\ToolEvolutionRightBox+\dp\ToolEvolutionRightBox-\ToolEvolutionPanelHeight)/5\relax}
  \setlength{\ToolEvolutionPanelHeight}{\dimexpr\ht\ToolEvolutionRightBox+\dp\ToolEvolutionRightBox\relax}
\fi
\typeout{Tool evolution final panel height: \the\ToolEvolutionPanelHeight; row padding: \the\ToolEvolutionRowPad}

% Typeset the final panels in the same measured context, then place their
% saved boxes on one baseline so both top and bottom edges align exactly.
\begin{lrbox}{\ToolEvolutionLeftBox}
\begin{minipage}{\ToolEvolutionLeftWidth}
\ToolEvolutionTable
\end{minipage}
\end{lrbox}
\begin{lrbox}{\ToolEvolutionRightBox}
\begin{minipage}{\ToolEvolutionRightWidth}
\ToolEvolutionRightPanel{height=\ToolEvolutionPanelHeight}
\end{minipage}
\end{lrbox}
\typeout{Tool evolution rendered left: \the\dimexpr\ht\ToolEvolutionLeftBox+\dp\ToolEvolutionLeftBox\relax}
\typeout{Tool evolution rendered right: \the\dimexpr\ht\ToolEvolutionRightBox+\dp\ToolEvolutionRightBox\relax}

\noindent
\begin{minipage}{\ToolEvolutionLeftWidth}
\centering
\fontsize{8}{9.2}\selectfont
\textbf{(a) Representative evolution outcomes}
\end{minipage}%
\hfill
\begin{minipage}{\ToolEvolutionRightWidth}
\centering
\fontsize{8}{9.2}\selectfont
\textbf{(b) Accepted executable tool (Round 2)}
\end{minipage}
\par\vspace{4pt}\nointerlineskip
\noindent\usebox{\ToolEvolutionLeftBox}\hfill\usebox{\ToolEvolutionRightBox}\par
\ToolTableStyleEnd
\end{table}

\subsection{Main Results}
\paragraph{Comparison with self-improving agents.}
VideoResearcher leads the evaluated self-improving agents on all three Overall scores and their average (Table~\ref{tab:main-results}).
VideoResearcher outperforms SkillOpt, Meta-Harness, and AutoResearch by 4.7, 1.0, and 2.8 points in average accuracy, respectively, with notable gains including 8.5 points over SkillOpt on LongVideoBench and 5.0 points over AutoResearch on Video-MME.
Compared with META, it improves Overall accuracy by 11.4 points on LongVideoBench and 2.5 points on Video-MME.
These results are consistent with the optimization scope: SkillOpt refines instructions and lowers Base accuracy by 2.3 points; the evaluated Meta-Harness and AutoResearch runs mainly revise prompts, yielding limited gains.
VideoResearcher adds two executable tools for better evidence acquisition (Table~\ref{tab:tool-evolution}): \emph{Neutral Scene Landmark Index} indexes coarse video events, and \emph{Candidate Window Atomic Scan} inspects short clips near candidate events.
Together, they raise Base's average accuracy from 72.1 to 74.5.

\paragraph{Comparison with human-engineered agents.}
Taking human-designed agents as the performance upper bound, VideoResearcher reaches this bound on LongVideoBench and Video-MME, leading in Overall accuracy, and ties for second on LVBench among the systems in Table~\ref{tab:main-results}.
Its autonomously developed tools combine coarse indexing with focused local inspection, echoing DVD~\citep{zhang2025dvd} and VideoSeek~\citep{lin2026videoseek}.
These parallels suggest that tool evolution recovers strategies from manual design, although the published configurations do not isolate its effect.\looseness=-1\par

\subsection{Ablation Studies}
\paragraph{Framework components.}
Agentic tool use and tool evolution both improve accuracy (Table~\ref{tab:ablation}(a)).
With Qwen3-VL-8B, adding the Worker raises LVBench accuracy from 58.0 to 63.6, and tool evolution raises it to 66.8, an 8.8-point gain over direct perception.
The complete framework reaches 68.4 with Qwen3.5-4B.
Both perception models benefit from combining these components.\par

\paragraph{Worker and perception models.}
Stronger Workers substantially improve accuracy, consistent with prior findings~\citep{zhang2025dvd,chen2026memdreamer}.
Table~\ref{tab:ablation}(c) tests three Workers (GPT-o4-mini, GPT-5.3-Codex, and GPT-5.6-SOL) with two perception models using the same evolved toolbox, without further evolution.
Changing the perception model yields smaller gains: with GPT-5.6-SOL, switching from Qwen3-VL-8B to Qwen3.5-4B raises accuracy from 66.8 to 68.4.\par

\paragraph{Effects of evolved tools.}
Further toolbox expansion may yield diminishing and uneven overall gains.
Table~\ref{tab:ablation}(b) shows gains over Base on LVBench, Video-MME, and LongVideoBench of 2.0, 3.7, and 2.3 points at Round 2, compared with 1.5, 3.9, and 1.8 points at Round 7, respectively.
Table~\ref{tab:ablation}(d) also supports the tools' intended function: Round 2 introduces the landmark index, raising accuracy in locating when events occur (temporal grounding, TG) from 67.7 to 71.4 (+3.7 points).

\begin{figure}[t]
  \centering
  \includegraphics[width=\linewidth]{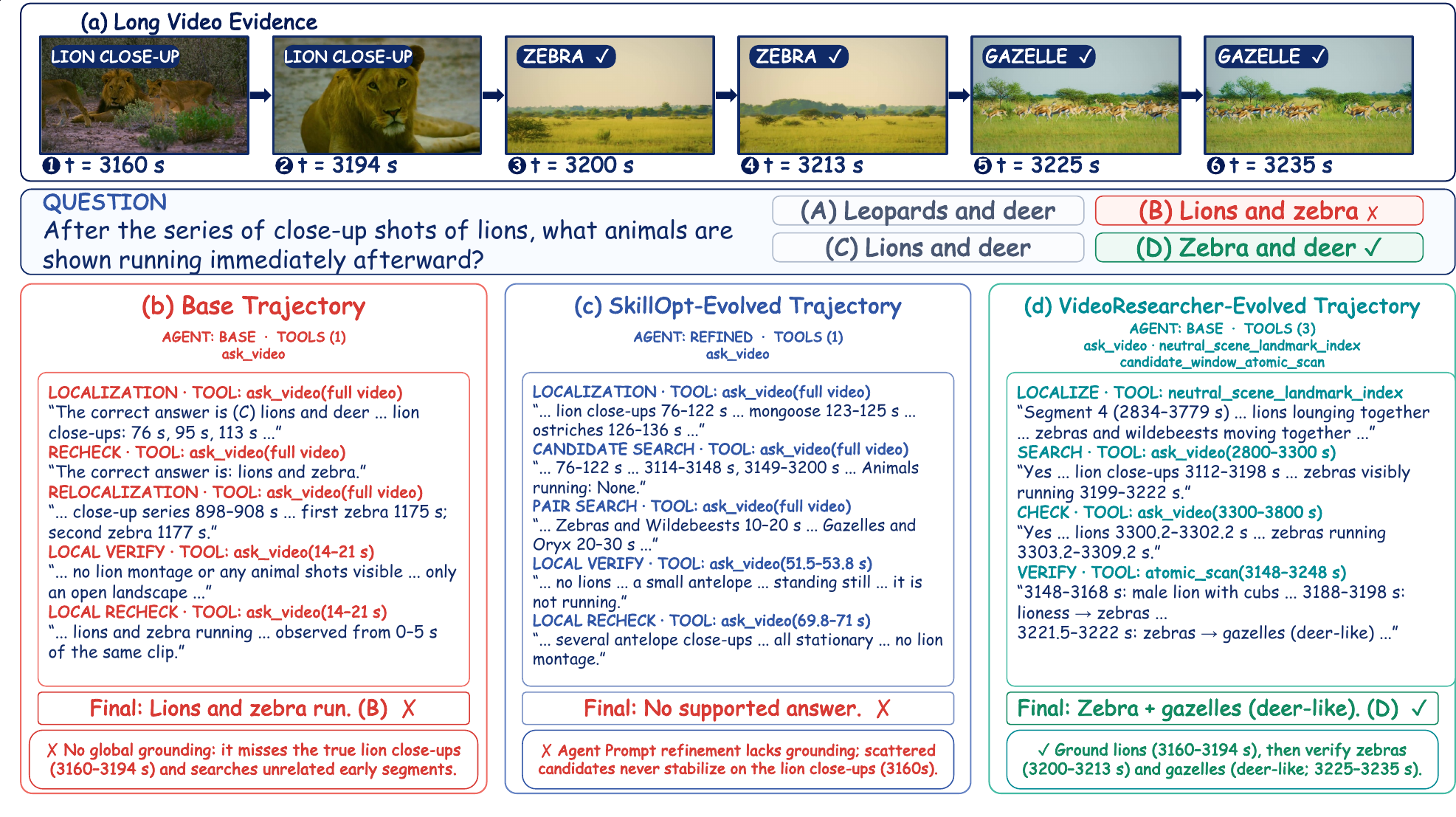}
  \caption{Tool-use trajectories for the same video and question. (a) Shared long-video evidence. (b) The base Worker uses the default \texttt{ask\_video} tool. (c) SkillOpt revises the Worker prompt with the same tool interface. (d) VideoResearcher uses evolved executable tools to retrieve decisive evidence and answer correctly.}
  \label{fig:qualitative-trajectory}
\end{figure}

\subsection{Qualitative Analysis of Tool Evolution}
\paragraph{Evolution outcomes.}
VideoResearcher proposes distinct tool behaviors for different evidence failures.
Table~\ref{tab:tool-evolution}(a) covers coarse indexing, action anchoring, local event scanning, cross-scene association, and context consolidation.
The landmark index and atomic scan are accepted, raising development accuracy from 68.7 to 75.0 and then 76.7; the other three proposals reduce accuracy and are rejected.
Panel~(b) shows the Round-2 tool's contract and executable implementation.
Thus, accepted tools expand the toolbox, while rejected proposals preserve unresolved problems and unsuccessful implementations for later evolution or manual design.\par

\paragraph{Trajectory case study.}
Figure~\ref{fig:qualitative-trajectory} illustrates how a new tool changes acquired evidence and the final answer.
Panel~(a) fixes the video and question; Panels~(b)--(d) compare Base, SkillOpt, and VideoResearcher.
Base misses the decisive transition with \texttt{ask\_video}; SkillOpt's revised Worker prompt retains the tool interface and still fails to retrieve sufficient evidence.
The evolved indexing and local-scanning tools locate the relevant interval, recover the zebra-and-gazelle evidence, and enable VideoResearcher to answer correctly.
This case connects an identified failure to a new capability, better evidence, and a corrected answer.

\FloatBarrier

\section{Conclusion}

In this work, we introduced VideoResearcher, a training-free multi-agent framework that automates tool development for long-video understanding.
By diagnosing evidence gaps in execution trajectories, developing executable tools, and validating accuracy gains, VideoResearcher turns observed failures into tested toolbox updates.
This process strengthens evidence acquisition while keeping the Worker and perception model fixed, allowing agents to expand their capabilities without repeated manual tool design.
Experiments on three benchmarks demonstrate state-of-the-art performance among self-improving agents, approaching the human-designed upper bound and raising average accuracy from 72.1\% to 74.5\%.

These results highlight autonomous tool development as a practical route to more capable long-video agents.
However, the current framework relies on textual trajectory assessment and repeated selection on one development set, and its evolution remains confined to tools.
Future work will investigate independent candidate-selection data and extend evolution to agent architectures, enabling adaptation in both evidence acquisition and reasoning organization.

\FloatBarrier

\bibliographystyle{abbrvnat}
\bibliography{paper}

\end{document}